\documentclass[twoside,11pt]{article}

\usepackage{blindtext}
\usepackage{float}
\usepackage{amsmath}
\usepackage{algorithm}
\usepackage{algpseudocode}
\usepackage{threeparttable}
\usepackage{booktabs}
\usepackage{subcaption}
\usepackage[preprint]{jmlr2e}

\usepackage{lastpage}

\begin{document}

\title{Proximal Policy Optimization with Evolutionary Mutations}

\author{
  \name Casimir Czworkowski \email cczwork1@jh.edu \\
  \addr Johns Hopkins University
  \AND
  \name Stephen Hornish \email shornis1@jh.edu \\
  \addr Johns Hopkins University
  \AND
  \name Alhassan S. Yasin \email ayasin1@jhu.edu \\
  \addr Johns Hopkins University
}
\maketitle

\begin{abstract}
Proximal Policy Optimization (PPO) is a widely used reinforcement learning algorithm known for its stability and sample efficiency, but it often suffers from premature convergence due to limited exploration. In this paper, we propose POEM (Proximal Policy Optimization with Evolutionary Mutations), a novel modification to PPO that introduces an adaptive exploration mechanism inspired by evolutionary algorithms. POEM enhances policy diversity by monitoring the Kullback–Leibler (KL) divergence between the current policy and a moving average of previous policies. When policy changes become minimal, indicating stagnation, POEM triggers an adaptive mutation of policy parameters to promote exploration. We evaluate POEM on four OpenAI Gym environments: CarRacing, MountainCar, BipedalWalker, and LunarLander. Through extensive fine-tuning using Bayesian optimization techniques and statistical testing using Welch’s t-test, we find that POEM significantly outperforms PPO on three of the four tasks (BipedalWalker: $t=-2.0642$, $p=0.0495$; CarRacing: $t=-6.3987$, $p=0.0002$; MountainCar: $t=-6.2431$, $p<0.0001$), while performance on LunarLander is not statistically significant ($t=-1.8707$, $p=0.0778$). Our results highlight the potential of integrating evolutionary principles into policy gradient methods to overcome exploration-exploitation tradeoffs.
\end{abstract}

\section{Introduction}
Reinforcement Learning (RL) algorithms often struggle with balancing exploration and exploitation. Models like Proximal Policy Optimization (PPO) tend to perform better when learning with small, incremental policy updates. These tiny updates have the negative side effect of potentially causing these models to converge prematurely in local optima, especially in environments with an ineffective reward structure. Standard PPO typically utilizes a Kullback–Leibler (KL) divergence-based policy update to prevent excessive policy changes, but that can unintentionally reduce exploration by discouraging the policy from deviating too far from what it already knows. The result is a sort of policy stagnation, where the model essentially learns a suboptimal policy instead of actively exploring different behaviors. 

In this paper, we propose a novel adaptation to PPO that incorporates an adaptive exploration mechanism heavily inspired by evolutionary algorithms. The core mechanism of the adaptation is to monitor the policy’s behavioral change over time using the KL divergence. Specifically, the new algorithm tracks the moving average of past policies to compute the KL divergence between the current policy and past reference policies. This effectively becomes a measure of policy novelty, where a high value quantitatively indicates that the policy is substantially different from its recent history. When the value becomes too low, it indicates that the policy is becoming too predictable or potentially stagnant, triggering an evolutionary mutation of the policy parameters. This way, if the algorithm hasn’t changed significantly in a while, it tries something different. These changes are random permutations in policy parameters, analogous to the mutations seen in genetic algorithms. The mutation is adaptive, scaling according to how low the diversity metric goes. After mutation, the model resumes standard PPO operations. 

We hypothesize that integrating a KL divergence-based diversity trigger with adaptive evolutionary mutations into PPO will significantly improve its performance on control tasks compared to standard PPO.

\section{Related Work}
Our PPO adaptation, which focuses primarily on monitoring policy diversity with KL divergence to trigger mutations, differs from other existing reinforcement learning approaches. Several papers suggest that diversity is a valuable metric worth exploring. For example, \citep{wu2023qualitysimilar} describe a method capable of learning diverse skills in an unsupervised manner by maximizing mutual information. Unlike our supervised, reward-focused policy approach, they aim to generate a repertoire of skills instead of just a single PPO policy, in order to enhance exploration. 

Another example is the gradient-based exploration described in the GPO model as shown in \citep{GPO}. This model uses gradient diversity to enhance exploration, but instead of utilizing a single KL divergence measure to trigger corrections, it adjusts continuously. Adjustments are made through gradient-based regularization. This method of continuously perturbing the model is also explored by \citep{Fortuno}, who inject a constant stream of noise into neural network parameters during training. Interestingly, the noise is sampled from a learned distribution. 

\citep{Gangwani} took a different view of model merging by incorporating PPO into existing evolutionary methods, a direction opposite to ours. Their work essentially uses PPO as a subroutine within an evolutionary model that focuses on evolving a population of diverse policies without any extrinsic rewards. The goal of their method is to produce a wide range of evolutionary behaviors. In contrast, our method strives to refine a single policy for a reward-driven task. 

\section{Methods}
Our POEM algorithm, detailed in Algorithm 1, extends standard PPO by integrating an adaptive exploration mechanism that leverages a moving average of past policy parameters. Unlike PPO, which relies solely on a clipped surrogate objective along with entropy regularization, POEM monitors the KL divergence between the current policy and its moving average to quantify policy novelty. When this divergence falls below a predefined threshold, an evolutionary mutation is triggered, whereby noise is adaptively injected into the policy parameters to escape local optima and encourage broader exploration. This additional mechanism distinguishes POEM from traditional PPO, providing a systematic way to overcome policy stagnation and achieve more robust performance.  

To properly evaluate the strength of our algorithm on a level playing field with PPO, we benchmarked on four OpenAI Gym environments: CarRacing (v0), MountainCar (v0), BipedalWalker (v3), and LunarLander (v2).

The CarRacing environment, based on the Box2D physics engine, used a 96×96 pixel RGB observation centered on the car. We converted frames to grayscale and stacked four frames at each step to allow velocity estimation. The action space was continuous with three actions: steering, acceleration (gas), and braking. We adopted the reward function
\[
  \text{Reward} = -0.1 \times (\text{Number of frames}) + \frac{1000}{\text{Number of track tiles visited}},
\]
which penalized slower completion and rewarded efficient track coverage.

MountainCar and BipedalWalker were used with their default settings and reward structures: MountainCar’s sparse per‐step penalty (–1 per timestep until reaching the flag) and BipedalWalker’s continuous forward‐progress and energy‐use reward. For LunarLander, we modified the default environment by introducing a fuel mechanic to ensure each episode ended with either a safe landing or a crash. The fuel system was initialized with a total of 600 units, where firing the main engine consumed 3 fuel units per timestep, and firing each side engine consumed 1 fuel unit per timestep. This modification enforced resource-aware decision-making throughout each landing attempt.

The reinforcement learning PPO model was built using the open‐source Stable‐Baselines3 (SB3) framework and modified with the PyTorch library to implement our adaptive exploration mechanism. As shown in Algorithm~1, POEM augments the standard PPO update with an adaptive KL-based mutation step.  We maintain an exponential moving average of the policy parameters
\[
  \hat\theta \;\leftarrow\;\beta\,\hat\theta + (1-\beta)\,\theta,
\]
where \(\theta\) are the current policy weights and \(\beta\in[0,1]\) controls smoothing.  On each minibatch we first compute the usual PPO clipped surrogate loss \(L_{\rm PPO}(\theta)\), plus the value-function loss \(L_{\rm VF}(\theta)\) and an entropy bonus \(H(\pi_\theta)\).  We then form the total loss
\[
  L_{\rm total}(\theta)
    = L_{\rm PPO}(\theta)
    - \lambda_{\rm div}\,D_{\rm KL}\bigl(\pi_\theta \,\|\, \pi_{\hat\theta}\bigr)
    + \alpha_{\rm vf}\,L_{\rm VF}(\theta)
    - \alpha_{\rm ent}\,H(\pi_\theta),
\]
where \(D_{\rm KL}(\pi\|\pi')\) is the average KL divergence over the minibatch, \(\lambda_{\rm div}\)  weights our diversity bonus, and \(\alpha_{\rm vf},\alpha_{\rm ent}\) are the value- and entropy-coefficients inherited from PPO.  After applying the gradient step, we recompute
\[
  d_{\rm post} \;=\; D_{\rm KL}\bigl(\pi_{\theta_{\rm new}}\|\pi_{\hat\theta}\bigr).
\]
If \(d_{\rm post}<\delta\), we set
\[
  \sigma \;=\;\sigma_{\min} + (\sigma_{\max}-\sigma_{\min})\,\frac{\delta - d_{\rm post}}{\delta},
\]
where $\delta$ is the divergence threshold and $\sigma$ is the adaptively interpolated standard deviation, clamped to \([\sigma_{\min},\sigma_{\max}]\), and sample one perturbed candidate
\(\theta' = \theta_{\rm new} + \mathcal{N}(0,\sigma^2 I)\). We then evaluate \(L_{\rm total}(\theta')\) and replace \(\theta_{\rm new}\!\leftarrow\!\theta'\) if and only if \(L_{\rm total}(\theta')<L_{\rm total}(\theta_{\rm new})\).  This mutation step (line 10–17 of Algorithm 1) enables POEM to escape local optima when standard PPO updates become too conservative.

\begin{algorithm}[H]
\caption{PPO with Evolutionary Mutation}\label{alg:poem} 
\begin{algorithmic}[1]
\State \textbf{Initialize} policy parameters $\theta$
\State \textbf{Initialize} variance of policy parameters $\sigma$
\State \textbf{Set} KL divergence threshold $\delta$
\State \textbf{Set} minimum and maximum mutation variances $\sigma_{\min}, \sigma_{\max}$
\While{not converged}
    \State Collect actions using policy $\pi_\theta$
    \State Compute advantages $\hat{A}_t$ and rewards-to-go $\hat{R}_t$
    \State Update moving average policy $\hat{\pi}_\theta$
    \State Compute KL divergence:
    \[
       D_{\text{KL}} \;=\; \frac{1}{N}\sum_{i=1}^{N} \log \frac{\pi_\theta(a_i \mid s_i)}{\hat{\pi}_\theta(a_i \mid s_i)}
    \]
    \If{$D_{\text{KL}} < \delta$}
        \State $\displaystyle \sigma \gets \sigma_{\min} \;+\; (\sigma_{\max} - \sigma_{\min}) 
        \times \frac{\delta - D_{\text{KL}}}{\delta}$
        \State Generate candidate mutations:
        \[
          \theta'(j) = \theta + \mathcal{N}(0,\,\sigma^2)
        \]
        \State Evaluate combined objective:
        \[
          L_{\rm total}(\theta') = L_{\rm PPO}(\theta')
  - \lambda_{\rm div}\,D_{\rm KL}\bigl(\pi_{\theta'}\|\pi_{\hat\theta}\bigr)
        \]
        \State Select the best candidate mutation:
        \If{$L_{\rm total}(\theta'(j)) < L_{\rm total}(\theta)$}
          \State $\theta \gets \theta'(j)$
        \EndIf
        
        \State Update policy parameters $\theta \gets \theta^*$
    \Else
        \State Perform standard PPO update on $\theta$
    \EndIf
\EndWhile

\end{algorithmic}
\end{algorithm}

Hyperparameter tuning was performed using Optuna prior to the final training phase. The range of hyperparameters was selected based on the original PPO algorithm's optimal values, allowing deviations of up to ±10\%. Optuna utilizes a Tree-structured Parzen Estimator (TPE) algorithm to efficiently identify optimal hyperparameters without exhaustive grid searches. Both POEM and PPO underwent hyperparameter tuning using the same methodology across all environments. Each hyperparameter combination was evaluated through a brief training run of 100,000 timesteps followed by a 4-episode evaluation for CarRacing, BipedalWalker, and LunarLander, and a shorter 50,000-timestep run followed by a 4-episode evaluation for MountainCar. All runs employed fixed random seeds and deterministic execution to ensure reproducibility. For each environment, the hyperparameter set yielding the highest average evaluation reward was selected for final training.

With the best hyperparameters identified for each environment, the final training phase consisted of 150,000, 250,000, 500,000, and 1,500,000 timesteps for MountainCar, LunarLander, CarRacing, and BipedalWalker, respectively. After training, each model was evaluated over 15 deterministic episodes with random seeds, and total rewards were recorded and averaged. Final model checkpoints and complete evaluation logs, including rewards formatted as CSV files, were saved to ensure reproducibility.

Throughout both the tuning and final training phases, TensorBoard logs provided real‐time monitoring, and performance plots were generated to visualize training dynamics and evaluation outcomes. This systematic, reproducible approach enabled a fair comparison between POEM and PPO across diverse control tasks.

\section{Experimental Results}
We ran 10 experiments each for PPO and POEM across four OpenAI Gym environments: CarRacing, MountainCar, BipedalWalker, and LunarLander, with each run initialized by a fixed random seed and executed deterministically to ensure reproducibility. The results of Welch’s two-sample t-tests comparing POEM to PPO on each environment are presented in Table \ref{tab:ttest_results}. POEM demonstrates statistically significant improvements over PPO on BipedalWalker ($t=-2.0642$, $p=0.0495$), CarRacing ($t=-6.3987$, $p=0.0002$), and MountainCar ($t=-6.2431$, $p<0.0001$), while the difference on LunarLander ($t=-1.8707$, $p=0.0778$) is not significant.

\begin{table}[H]
\centering
\caption{Summary of t-test Results Comparing PPO vs. POEM}
\label{tab:ttest_results}
\begin{tabular}{lrrc}
\toprule
\textbf{Test Name} & \textbf{T-statistic} & \textbf{P-value} & \textbf{POEM Significantly Better?} \\
\midrule
Bipedal Walker     & -2.0642              & 0.0495           & Yes  \\
CarRacing          & -6.3987              & 0.0002           & Yes  \\
Mountain Car           & -6.2431              & $<0.0001$           & Yes  \\
Lunar Lander       & -1.8707              & 0.0778           & No   \\
\bottomrule
\end{tabular}
\end{table}

\begin{figure}[ht]
  \centering
  \begin{subfigure}[b]{0.47\textwidth}
    \centering
    \includegraphics[width=\textwidth]{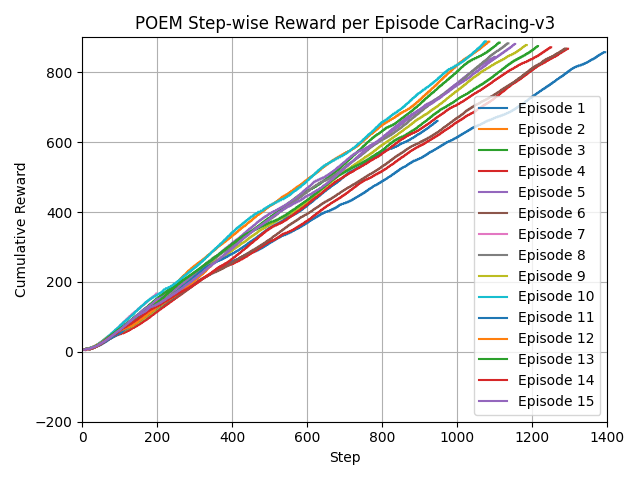}
    \caption{POEM CarRacing-v3}
    \label{fig:poem_carracing_reward}
  \end{subfigure}
  \hfill
  \begin{subfigure}[b]{0.47\textwidth}
    \centering
    \includegraphics[width=\textwidth]{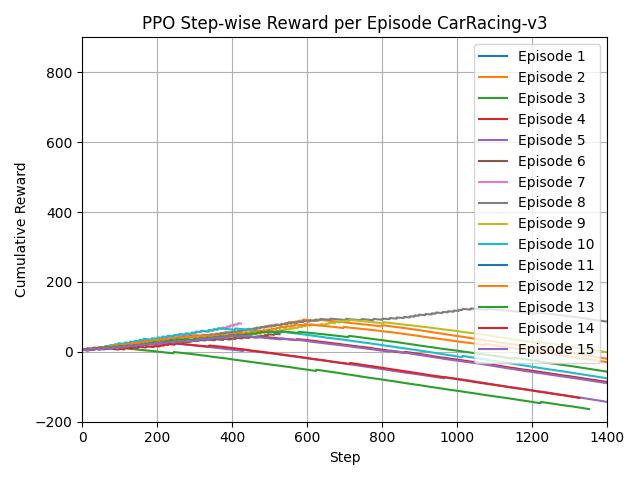}
    \caption{PPO CarRacing-v3}
    \label{fig:ppo_carracing_reward}
  \end{subfigure}

  \caption{
    Cumulative reward per step on CarRacing-v3 for POEM and PPO across 15 evaluation episodes.
    \textbf{(a) POEM:} shows low variance across seeds, reliably progresses through the track, and completes all but four runs.
    \textbf{(b) PPO:} frequently becomes stuck on turns and fails to complete any episode.
  }
  \label{fig:carracing_comparison}
\end{figure}

\begin{figure}[ht]
  \centering
  \begin{subfigure}[b]{0.47\textwidth}
    \centering
    \includegraphics[width=\textwidth]{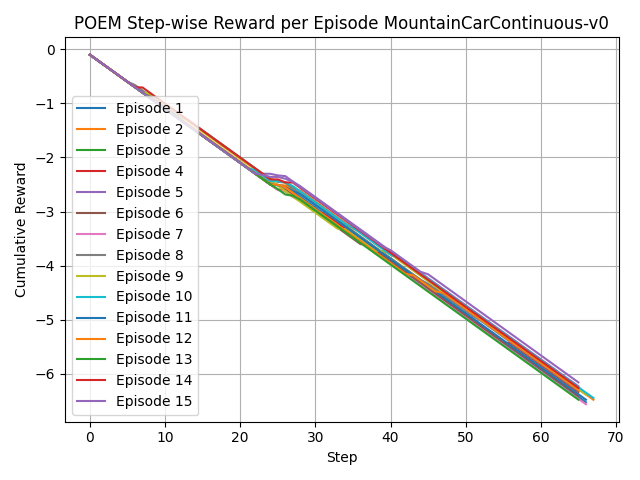}
    \caption{POEM MountainCarContinuous-v0}
    \label{fig:poem_mountaincar_reward}
  \end{subfigure}
  \hfill
  \begin{subfigure}[b]{0.47\textwidth}
    \centering
    \includegraphics[width=\textwidth]{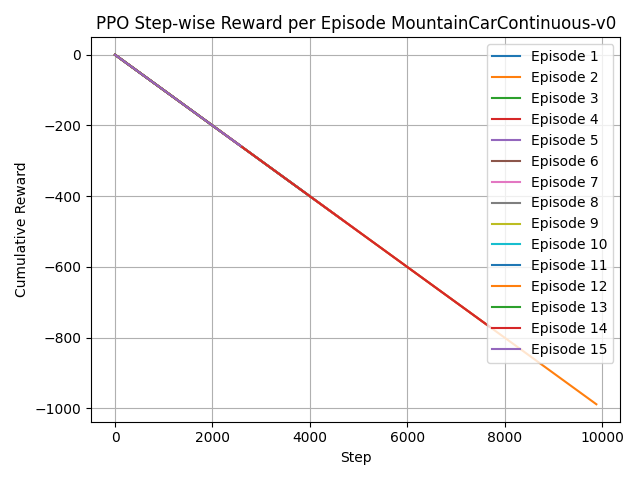}
    \caption{PPO MountainCarContinuous-v0}
    \label{fig:ppo_mountaincar_reward}
  \end{subfigure}

  \caption{
    Cumulative reward per step on MountainCarContinuous-v0 for POEM and PPO across 15 evaluation episodes .
    \textbf{(a) POEM:} successfully explores the environment, escapes the valley, and consistently reaches the goal incurring minimal negative reward.
    \textbf{(b) PPO:} fails to solve the task in all episodes, reaching the maximum timestep limit or exiting the bounds of the course and incurring the full 1000-step cumulative cost. Note Y scales are different between graphs and final reward of +100 for completing the task is not graphed for readability. 
  }
  \label{fig:mountaincar_comparison}
\end{figure}

\begin{figure}[ht]
  \centering
  \begin{subfigure}[b]{0.47\textwidth}
    \centering
    \includegraphics[width=\textwidth]{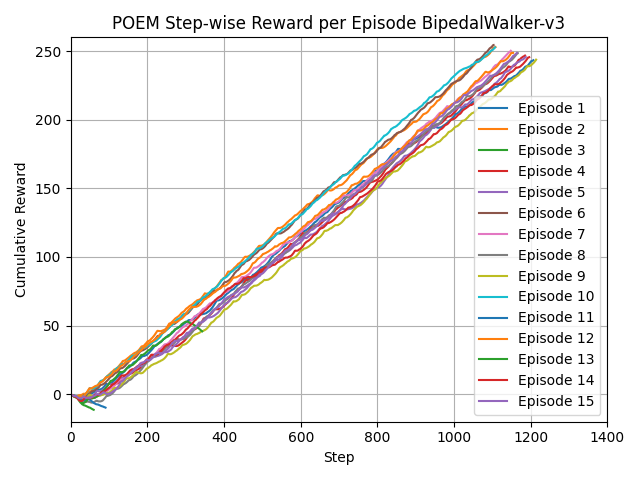}
    \caption{POEM}
    \label{fig:poem_bipedalwalker_reward}
  \end{subfigure}
  \hfill
  \begin{subfigure}[b]{0.47\textwidth}
    \centering
    \includegraphics[width=\textwidth]{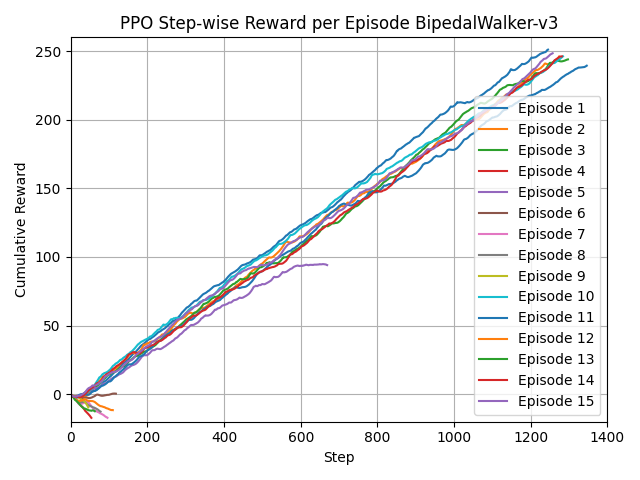}
    \caption{PPO}
    \label{fig:ppo_bipedalwalker_reward}
  \end{subfigure}

  \caption{
    Cumulative reward per step for POEM and PPO on BipedalWalker-v3 across 15 evaluation episodes. 
    \textbf{(a) POEM:} reward trajectories show low variance across seeds, indicating stable performance. POEM reliably progresses across the terrain, with all but three episodes completing the course, and all successful episodes finishing in under 1200 steps.
    \textbf{(b) PPO:} completes the course in only 7 out of 15 episodes, with the remainder resulting in failure. No PPO run completes the course in fewer than 1200 steps.
  }
  \label{fig:bipedalwalker_comparison}
\end{figure}

\begin{figure}[ht]
  \centering
  \begin{subfigure}[b]{0.47\textwidth}
    \centering
    \includegraphics[width=\textwidth]{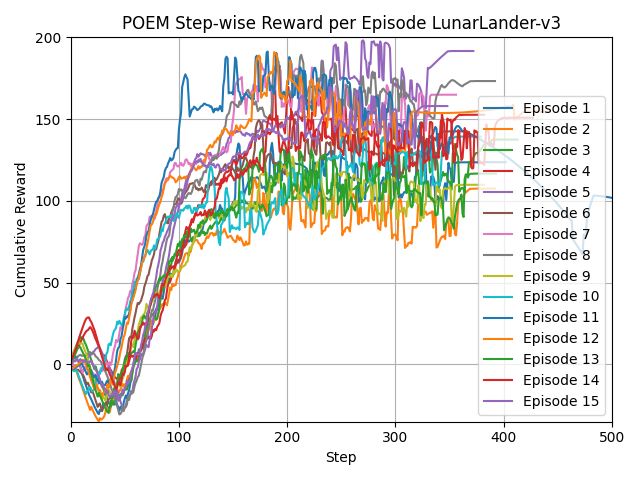}
    \caption{POEM}
    \label{fig:poem_lunarlander_reward}
  \end{subfigure}
  \hfill
  \begin{subfigure}[b]{0.47\textwidth}
    \centering
    \includegraphics[width=\textwidth]{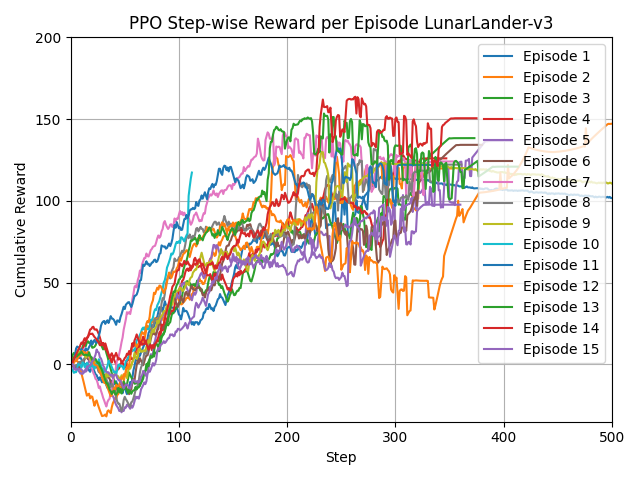}
    \caption{PPO}
    \label{fig:ppo_lunarlander_reward}
  \end{subfigure}
  \caption{
    Cumulative reward per step for POEM and PPO on LunarLander-v3 across 15 evaluation episodes.
    \textbf{(a) POEM:} reward increases rapidly during the first 100 steps as the agent stabilizes its descent, followed by a plateau where the lander fine-tunes its position before performing a controlled landing.
    \textbf{(b) PPO:} exhibits a more gradual reward increase, reflecting a strategy that involves searching for a landing region at higher altitude before descending and attempting the final landing maneuver.
  }
  \label{fig:lunarlander_comparison}
\end{figure}

Figures \ref{fig:carracing_comparison}–\ref{fig:lunarlander_comparison} show the step-wise cumulative reward for both PPO and POEM across the four test environments. Each plot reports results from 15 fixed-seed evaluation episodes. Across all environments, POEM accumulates reward more quickly and consistently achieves higher final returns than PPO.

\clearpage

\begin{figure}[H]
  \centering
  \begin{subfigure}[b]{0.45\textwidth}
    \includegraphics[width=\textwidth]{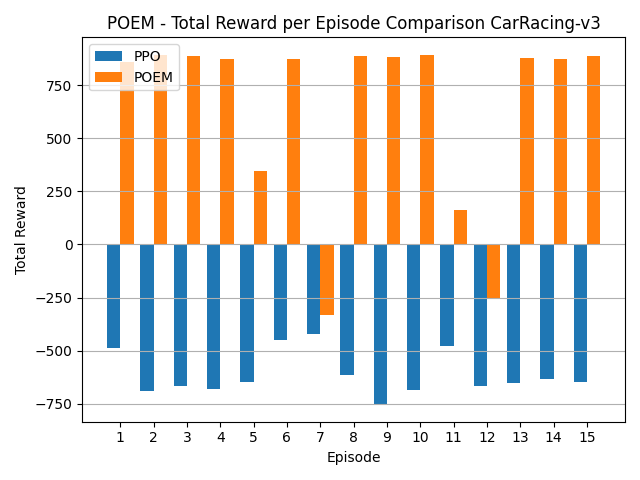}
    \caption{CarRacing}
    \label{fig:bar_carracing}
  \end{subfigure}
  \hfill
  \begin{subfigure}[b]{0.45\textwidth}
    \includegraphics[width=\textwidth]{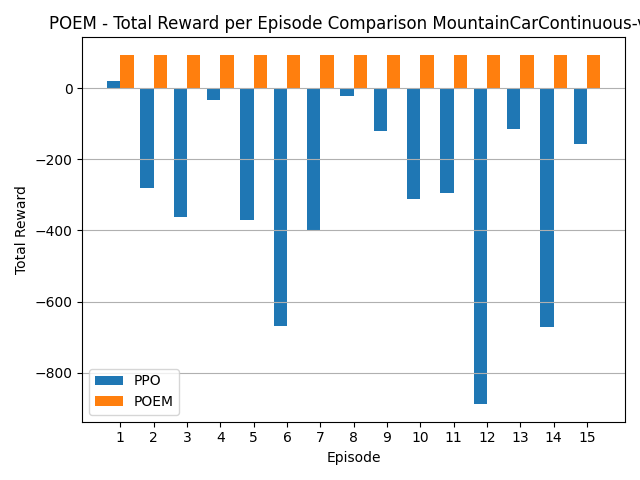}
    \caption{MountainCar}
    \label{fig:bar_mountaincar}
  \end{subfigure}

  \vspace{1em}

  \begin{subfigure}[b]{0.45\textwidth}
    \includegraphics[width=\textwidth]{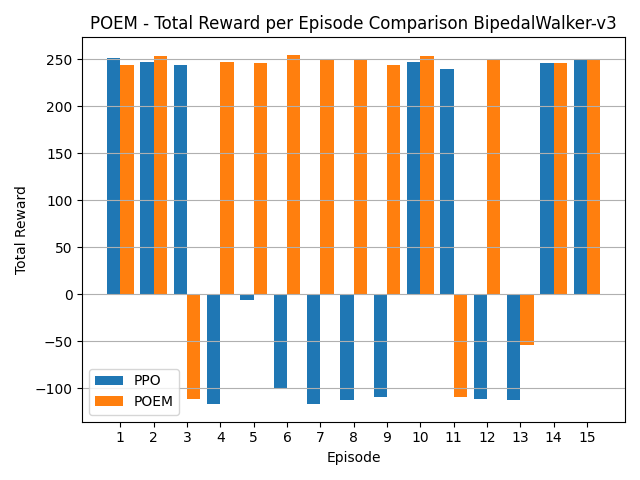}
    \caption{BipedalWalker}
    \label{fig:bar_bipedalwalker}
  \end{subfigure}
  \hfill
  \begin{subfigure}[b]{0.45\textwidth}
    \includegraphics[width=\textwidth]{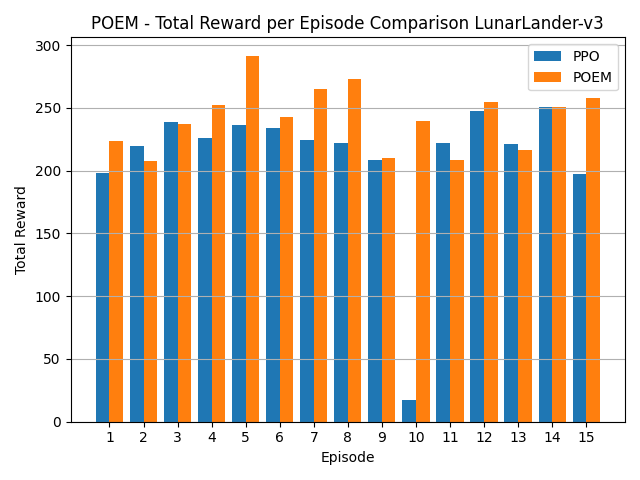}
    \caption{LunarLander}
    \label{fig:bar_lunarlander}
  \end{subfigure}

  \caption{Episode-wise total rewards for PPO (blue) and POEM (orange) across 15 runs. POEM wins the vast majority of runs; PPO only exceeds POEM in 1 episode of BipedalWalker and 4 episodes of LunarLander.}
  \label{fig:bar_episode_rewards}
\end{figure}

Figure \ref{fig:bar_episode_rewards} presents the total reward achieved in each of the 10 deterministic, randomly‐seeded runs (blue = PPO, orange = POEM). POEM outperforms PPO in almost every episode; PPO only narrowly beats POEM once in BipedalWalker and four times in LunarLander.

\begin{table}[h!]
\centering
\caption{Average Agent Reward per Environment over 15 Evaluation Episodes}
\label{tab:avg_reward}
\begin{tabular}{lcc}
\toprule
\textbf{Environment} & \textbf{POEM} & \textbf{PPO} \\
\midrule
CarRacing-v3                 & 640.01  & -610.83 \\
MountainCarContinuous-v0     & 93.52   & -311.75 \\
BipedalWalker-v3             & 180.58  & 62.43 \\
LunarLander-v3               & 242.10  & 210.94 \\
\bottomrule
\end{tabular}
\end{table}

\section{Discussion}
The experimental results across four OpenAI Gym environments---CarRacing, MountainCar, 
BipedalWalker, and LunarLander---demonstrate that POEM outperforms PPO in three of the 
four tasks when evaluated over ten deterministic, randomly seeded runs. Using Welch’s 
t-tests with a significance threshold of $\alpha = 0.05$, we find significant improvements 
for POEM on BipedalWalker ($t = -2.0642$, $p = 0.0495$), CarRacing ($t = -6.3987$, 
$p = 0.0002$), and MountainCar ($t = -6.2431$, $p < 0.0001$), while the difference on 
LunarLander ($t = -1.8707$, $p = 0.0778$) did not reach significance. The cumulative 
reward curves (Figures 1 - 4) and episode-reward 
bar charts (Figure~\ref{fig:bar_episode_rewards}) further illustrate POEM’s consistent superiority, 
with PPO only narrowly exceeding POEM once in BipedalWalker and four times in 
LunarLander.

\paragraph{CarRacing-v3.}
CarRacing exhibits the clearest performance gap between the two methods. As shown in 
Figure~\ref{fig:poem_carracing_reward}, POEM rapidly accumulates reward early in each episode and maintains stable control throughout, 
whereas PPO repeatedly collapses into low-reward trajectories (Figure~\ref{fig:ppo_carracing_reward}). The large 
t-statistic and extremely small $p$-value further underscore POEM’s decisive advantage, driven by its ability to avoid early-episode instability.

\paragraph{MountainCarContinuous-v0.}
MountainCar similarly favors POEM. PPO frequently fails to build sufficient momentum to 
escape the valley and earn the sparse reward, often stalling and timing out or taking a long time to build enough momentum to escape receiving large negative reward Figure~\ref{fig:ppo_mountaincar_reward}. POEM reliably reaches the 
goal and secures substantially higher returns shown by its small negative reward incurred during episodes and limited number of steps prior to episode completition as shown in Figure~\ref{fig:poem_mountaincar_reward}. The statistical result ($p < 0.0001$) and average mean rewards in Table~\ref{tab:avg_reward} reflects this consistent behavioral advantage.

\paragraph{BipedalWalker-v3.}
In BipedalWalker, both methods eventually learn viable movement, but POEM produces 
smoother and more stable trajectories, leading to higher average returns and success rate. 
PPO was inconsistent in its ability to cross the terrain, crashing in 8 out of 15 episodes. 
Even in the episodes where PPO succeeded, its completion time was still greater than that 
of the worst POEM run as shown in comparing Figure~\ref{fig:poem_bipedalwalker_reward} to Figure~\ref{fig:ppo_bipedalwalker_reward}.

\paragraph{LunarLander-v3.}
LunarLander presents a more nuanced comparison. The lack of statistical significance may 
stem from higher reward variance or residual hyperparameter sensitivity in this 
environment. When evaluating the stepwise reward in Figure~\ref{fig:poem_lunarlander_reward}, 
it is noteworthy that POEM adopted a different strategy than PPO. POEM favored a rapid 
early descent, gaining reward quickly and then plateauing as it fine-tuned its landing for 
terminal reward. PPO, in contrast, exhibited a gradual reward increase, hovering to find an 
optimal landing spot at high altitudes before committing to descent. However, as shown in 
Figure~\ref{fig:ppo_lunarlander_reward}, PPO’s hovering strategy produced one catastrophic 
failure and overall underperformed. This reflects an intrinsic risk; hovering at high altitude 
can lead to fuel exhaustion and a crash. POEM’s strategy of descending early and making 
corrections near the ground ensures that running out of fuel is less likely to result in 
total failure and also yielded a higher average reward.

Collectively, these findings suggest that POEM’s adaptive mutation mechanism effectively 
balances exploration and exploitation across diverse control tasks, enabling improved 
stability and performance in environments with both sparse and dense rewards.

\section{Conclusion}
In this work we introduced POEM, a Proximal Policy Optimization variant that injected evolutionary mutations whenever the KL divergence between the current and a moving-average policy fell below a threshold. Across 15 deterministic runs on four OpenAI Gym control tasks, POEM consistently achieved higher returns than vanilla PPO and delivered statistically significant gains on CarRacing, MountainCar, and BipedalWalker, while matching PPO on LunarLander.  These results confirmed our hypothesis that an adaptive, diversity-driven mutation step can unlock additional exploration without sacrificing PPO’s hallmark stability and sample efficiency.

Future work includes evaluating POEM's performance explicitly across environments with continuous and discrete action spaces to determine whether its advantages are action-space dependent. Additionally, extending POEM to more complex, high-dimensional environments could provide valuable insights into its scalability and robustness for real-world applications. Finally, although PPO remains widely used, future experiments should benchmark our modifications against other contemporary PPO variants. One example is PPO enhanced with intrinsic curiosity modules \citep{PPOICM}.

\vskip 0.2in
\bibliography{refs}

@article{article,
author = {Podgorelec, Vili and Kokol, Peter and Stiglic, Bruno and Rozman, Ivan},
year = {2002},
month = {11},
pages = {445-63},
title = {Decision Trees: An Overview and Their Use in Medicine},
volume = {26},
journal = {Journal of medical systems},
doi = {10.1023/A:1016409317640}
}

@inproceedings{
wu2023qualitysimilar,
title={Quality-Similar Diversity via Population Based Reinforcement Learning},
author={Shuang Wu and Jian Yao and Haobo Fu and Ye Tian and Chao Qian and Yaodong Yang and QIANG FU and Yang Wei},
booktitle={The Eleventh International Conference on Learning Representations },
year={2023},
url={https://openreview.net/forum?id=bLmSMXbqXr}
}

@misc{GPO,
      title={Policy Optimization by Genetic Distillation}, 
      author={Tanmay Gangwani and Jian Peng},
      year={2018},
      eprint={1711.01012},
      archivePrefix={arXiv},
      primaryClass={stat.ML},
      url={https://arxiv.org/abs/1711.01012}, 
}

@misc{Fortuno,
      title={Parameter Space Noise for Exploration}, 
      author={Matthias Plappert and Rein Houthooft and Prafulla Dhariwal and Szymon Sidor and Richard Y. Chen and Xi Chen and Tamim Asfour and Pieter Abbeel and Marcin Andrychowicz},
      year={2018},
      eprint={1706.01905},
      archivePrefix={arXiv},
      primaryClass={cs.LG},
      url={https://arxiv.org/abs/1706.01905}, 
}

@misc{Gangwani,
      title={Diversity is All You Need: Learning Skills without a Reward Function}, 
      author={Benjamin Eysenbach and Abhishek Gupta and Julian Ibarz and Sergey Levine},
      year={2018},
      eprint={1802.06070},
      archivePrefix={arXiv},
      primaryClass={cs.AI},
      url={https://arxiv.org/abs/1802.06070}, 
}

@article{PPOICM,
title = {Deep reinforcement learning with intrinsic curiosity module based trajectory tracking control for USV},
journal = {Ocean Engineering},
volume = {308},
pages = {118342},
year = {2024},
issn = {0029-8018},
doi = {https://doi.org/10.1016/j.oceaneng.2024.118342},
url = {https://www.sciencedirect.com/science/article/pii/S0029801824016809},
author = {Chuanbo Wu and Wanneng Yu and Weiqiang Liao and Yanghangcheng Ou}
}

\end{document}